\documentclass[conference]{IEEEtran}
\usepackage[pdftex]{graphicx}
\graphicspath{{images/}}
\DeclareGraphicsExtensions{.pdf,.jpeg,.png, .jpg}
\usepackage[ruled, lined, linesnumbered, commentsnumbered, longend]{algorithm2e}
\usepackage[cmex10]{amsmath}

\usepackage{array}
\usepackage{mdwmath}
\usepackage{mdwtab}
\usepackage{hyperref}
\hypersetup{colorlinks,allcolors=black}
\usepackage{eqparbox}
\usepackage[caption=false, font=footnotesize]{subfig}
\usepackage{url}

\begin{document}
%
\title{Indoor Semantic Scene Understanding using Multi-modality Fusion}




\author{\IEEEauthorblockN{Muraleekrishna Gopinathan\IEEEauthorrefmark{1},
Giang Truong\IEEEauthorrefmark{2},
Jumana Abu-Khalaf\IEEEauthorrefmark{3}}
\IEEEauthorblockA{School of Science,
Edith Cowan University,
Western Australia, Australia 6027\\ 
Email: \IEEEauthorrefmark{1}k.gopinathan@ecu.edu.au,\IEEEauthorrefmark{2}hagiangt@our.ecu.edu.au,\IEEEauthorrefmark{3}j.abukhalaf@ecu.edu.au}}


\maketitle

\begin{abstract}
Seamless Human-Robot Interaction is the ultimate goal of developing service robotic systems. For this, the robotic agents have to understand their surroundings to better complete a given task. Semantic scene understanding allows a robotic agent to extract semantic knowledge about the objects in the environment. In this work, we present a semantic scene understanding pipeline that fuses 2D and 3D detection branches to generate a semantic map of the environment. The 2D mask proposals from state-of-the-art 2D detectors are inverse-projected to the 3D space and combined with 3D detections from point segmentation networks. Unlike previous works that were evaluated on collected datasets, we test our pipeline\footnote{Code available at: \url{https://github.com/gmuraleekrishna/Benchbot-SSLAM}} on an active photo-realistic robotic environment \textit{BenchBot}. Our novelty includes rectification of 3D proposals using projected 2D detections and modality fusion based on object size. This work is done as part of the Robotic Vision Scene Understanding Challenge (RVSU). The performance evaluation demonstrates that our pipeline has improved on baseline methods without significant computational bottleneck.
\end{abstract}


%

\section{Introduction}
Human-Robot collaboration requires humans and robots to perform tasks by sharing workspaces and resources to achieve a goal effectively \cite{Das2020}. To meet this requirement, the robotic agent should be able to comprehend the environment and take appropriate actions based on the task at hand. In these scenarios, scene understanding is a critical aspect where the agent learns to perceive and analyse the environment. A typical pipeline comprises of object detection and scene graph (object-to-object and object-to-scene relationships) generation. In particular, scene understanding requires the robot agent to discover and detect all objects and their poses, based on the signals from sensors, then add them into a semantic map.

Although scene understanding is a trivial task for humans, it is still challenging for robots. In the recent years, due to the rise of deep learning-based models and proliferation of large datasets, 2D scene understanding has attracted a lot of attention from the research community. Some methods have been proposed to achieve reasonable accuracy on 2D scene understanding \cite{Shen2021}. However, there still is a gap in performance and generalisability on 3D scene understanding tasks, compared to their 2D counterparts.

Many robots are equipped with a wide variety of sensors, such as RGBD cameras, IMUs, etc., to comprehend the environment reliably. Moreover, the emergence of new embodied scene understanding tasks \cite{Anderson2018, Ku2020, Zhu2021b} have underscored the importance of cross-modal learning between different modalities including RGB image, depth image and language instructions. These methods rely on 2D object and place recognition models to generate semantic signals for the cross-modal matching using transformers \cite{Li2020e} and reinforcement learning \cite{Ku2020}. Visual modalities such as RGB and depth contribute the visual semantic cues to these methods. Hence improving semantic scene understanding can accelerate research in vision language navigation (VLN)  \cite{Anderson2018} , visual question answering (VQA) \cite{Antol2015}, rearrangement \cite{Batra2020} and visual commonsense reasoning \cite{Zellers2019} tasks.

\begin{figure*}[ht]
    \centering
    \includegraphics[width=\textwidth]{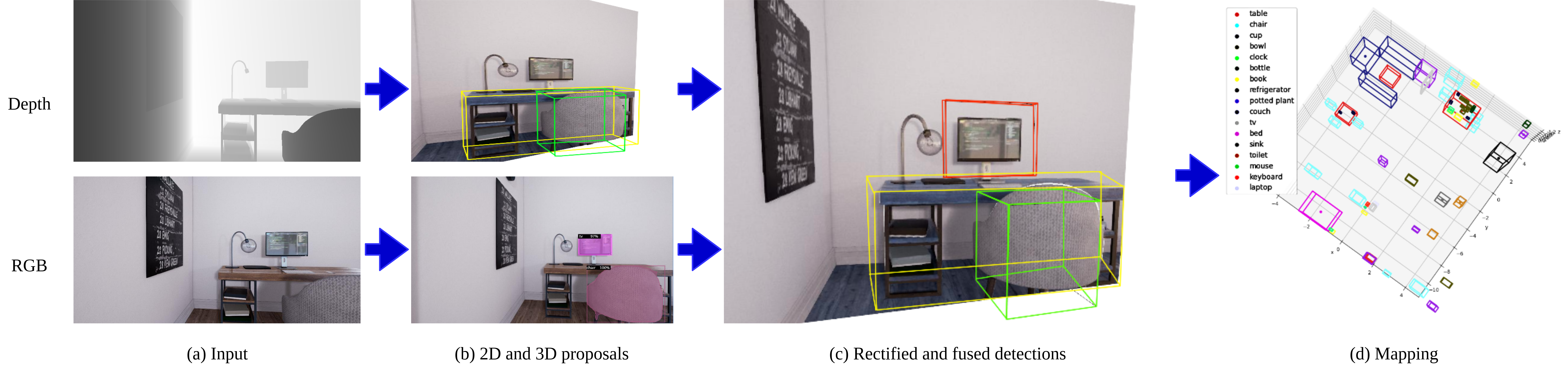}
    \caption{Illustration of the semantic scene understanding task: (a) input RGB-D image from the simulator (b) 2D and 3D independent proposals (c) Proposal Fusion, 3D proposals are rectified using 2D detections (d) Transforming detections to the global map}
\label{fig:ssu-task}
\end{figure*}

In this paper, we are interested in the problem of fusing multiple sensor modalities for semantic scene understanding. Fig.\ref{fig:ssu-task} illustrates an example visualization of our method. The robot agent moves around a household environment and captures a large amount of RGB-D images. Next, a detector is employed for 3D object detection. The output is then used to generate the object map. The detector is a critical element in our pipeline. Some methods \cite{Qi2019b, Qi2020a} convert RGB-D images to point clouds and directly predict 3D object detection from these point clouds. However, these methods have some limitations. First, the point clouds are usually sparse and have noise (due to depth sensor), the small objects may have just a few points and 3D detectors may miss them. Moreover, the imbalance in the dataset may lead to some undesired issues. Indeed, some objects only appear a few times in a training set. This scenario may negatively impact the learning process as well as the performance of the detector.
We propose novel techniques to improve the performance for semantic scene understanding. We found that 2D detectors, pre-trained with large datasets, can recognise small objects with high accuracy. Intuitively, we then leverage the output of a 2D detector to rectify the performance of a 3D detector. This also helps identify the false positive cases resulting from a 3D detector. In general, our contributions can be summarised as follows:
\begin{itemize}
    \item a detection pipeline which fuses 2D and 3D detection methods based on their performance on object classes.
    \item a proposal selection method that uses 2D detector proposals to remove false positive 3D detections.
    \item a scene understanding pipeline that can generate a semantic map from a photo-realistic simulator environment
\end{itemize}

 



\section{Related work}

\begin{figure*}[ht]
    \centering
    \subfloat[VoteNet detection]{%
        \includegraphics[width=0.32\textwidth]{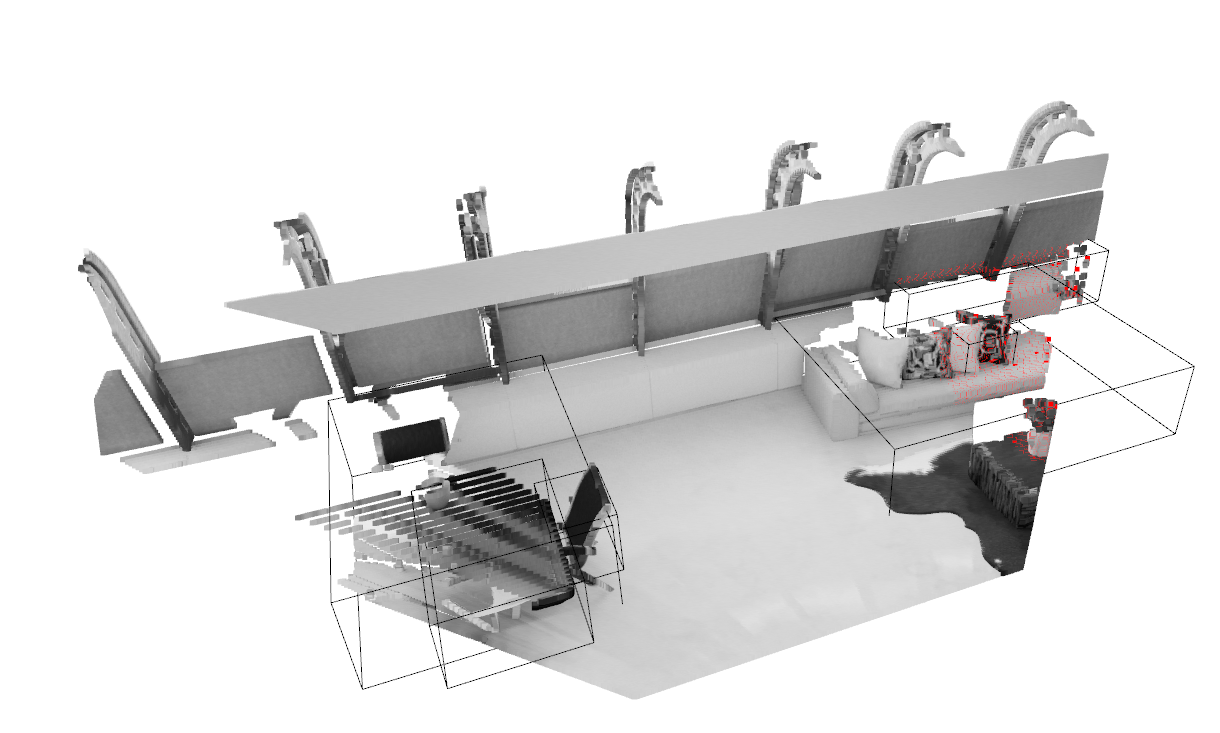}
        \label{fig:votenet_detection}
    }
    \subfloat[2D-3D re-projected proposal]{%
        \includegraphics[width=0.32\textwidth]{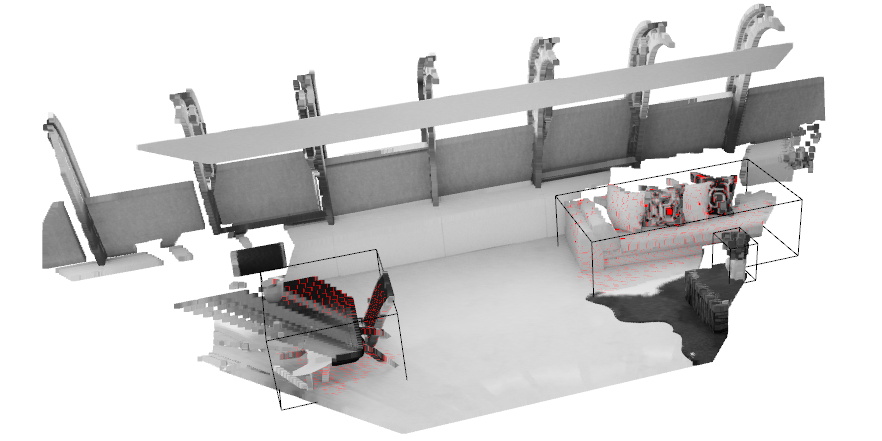}
        \label{fig:2d3dreprojected-proposal}
    }
    \subfloat[Duplicate and mis-oriented bounding boxes]{%
        \includegraphics[width=0.32\textwidth]{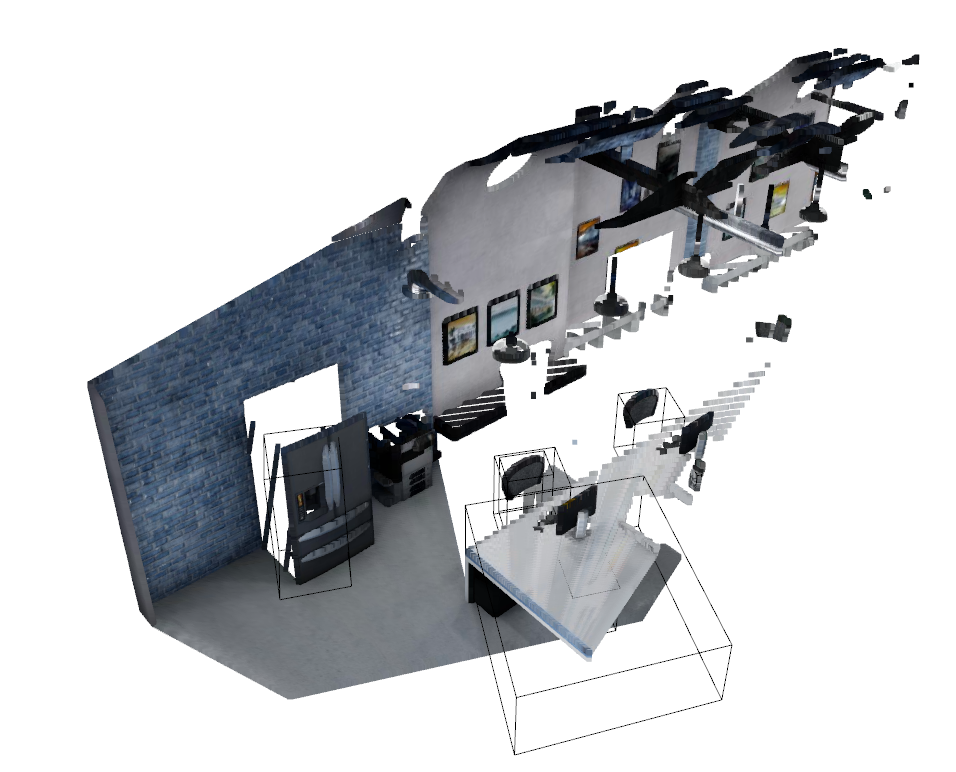}
        \label{fig:proposal-bunging}
    }
\caption{Illustration of problems with bounding box proposals from basic detection methods. (a) shows VoteNet detection overestimating the object boundaries for couch, table, and chair. (b) is the result of simple 2D-3D re-projection based on depth. Duplicate and incorrect pose for proposals after basic fusion. Our method overcomes these problems by using boundary preservation and shape completion strategies.}
\label{fig:basic-detection}
\end{figure*}

There is a large body of work in 2D \cite{Huang2019c} and 3D object detection \cite{Qi2020a}, instance, semantics, and panoptic segmentation \cite{Wang2020} using learning-based models. Instance level segmentation using Mask RCNN \cite{He2017} extends the state-of-the-art object detector Faster R-CNN with a Fully Connected Network (FCN) to produce an object segmentation mask. Boundary-Preserving Mask RCNN (BMask R-CNN) \cite{Cheng2020} improves this idea by retaining the mask boundaries via adding a boundary loss to the multi-task learning pipeline. 3D based detectors use CNN and Transformers to detect and segment objects directly from point clouds \cite{Qi2020a} or voxels \cite{Zhou2018} (volumetric elements). Recent works on 3D semantic instance reconstruction RfD-Net \cite{Nie2020} detects and reconstructs dense object surfaces for predicting shapes while infilling missing geometries caused by occlusion. Apart from the previous works on depth in-painting \cite{Huang2019d} and shape retrieval \cite{Avetisyan2019b}, RfD-Net was able to overcome the resolution issue by applying a 2-stage proposal and probabilistic shape learning approach. However, this method has poor performance on smaller objects with high point cloud sparsity. 

An evaluation of Semantic Scene Understanding (SSU) using an active agent was put forward by \cite{Hall2020b} and adopted in the Robotic Vision Scene Understanding (RVSU) challenge in the IEEE CVPR 2021 workshop. Unlike common metrics used in semantic SLAM challenges such as relative pose error (RPE) and absolute trajectory error (ATE) that test trajectory quality, their metric object map quality (OMQ) focuses on 4 parameters of object mapping namely pairwise, spatial, label, and false positive qualities. The OMQ score weighed on high spatial and label correctness of the detections and low false positive count. This paper specifically addresses the Semantic SLAM (SSLAM) sub-challenge, where the active agent has to explore various test environments and produce an object semantic map ($M$) with high OMQ score. Given the object cuboid $C_i = \{x_c, y_c, z_c, x_e, y_e, z_e\}$ with centroid $(x_c,y_c,z_c)$  and box extends  $(x_e, y_e, z_e)$ , and label probability ($l$) across all object classes of interest, the semantic map $M$ is given by $M = \{\{C_1, l_i\}, \{C_2, l_2\}, \{C_3, l_3\},..., \{C_n, l_n\}\}$. In the RVSU challenge \cite{Hall2020b}, the first baseline used VoteNet as the object detector producing a lower OMQ score. From our preliminary study, we found the VoteNet model used in their first baseline was over estimating the object sizes (Fig. \ref{fig:basic-detection}). The second baseline used a combination of Depth Segmentation \cite{Furrer2018} and BMask R-CNN \cite{Cheng2020} for 3D and 2D detections, respectively \cite{Hall2021}. While this method improved the OMQ score significantly, the model was unable to detect some partially occluded objects and large objects such as a conference table (Fig. \ref{fig:proposal-bunging}).  Additionally, the model expect larger objects to be placed on the ground and smaller objects to be above ground level. It also rectifies the small object using predefined bounding box sizes. Hence, this method failed to register large objects, such as a television, above the ground.

The objective of our proposed model is to improve performance over a wide range of object sizes. We propose to categorise objects by utilising the shape accuracy of 3D detection models for large objects, and consistent object mask accuracy of 2D detection models on all object sizes. Additionally, the model intends to use the class accuracy of 2D models to suppress false positive detections in the 3D model predictions.

\section{Background}
In this section we present the 3D detection and shape completion techniques evaluated in this work. We highlight the short comings of 3D-only methods and the motivation of using 2D detectors to support the 3D detection methods. The OMQ evaluation metric used by the RVSU challenge is also briefly introduced.
\subsection{3D detection and segmentation}
VoteNet \cite{Qi2019b}  is a 3D detection method that predicts 3d oriented bounding boxes directly from point clouds using Deep Hough Voting. The backbone network, PointNet++ \cite{Qi2017}, generates M seed points $\{s_i\}_{i=1}^M$ where each point is $s_i = [x_i, f_i]$ with position $x_i \in R^3$ and feature vector $f_i \in R^C$. A multi-layer perceptron (MLP) with fully connected layers is used to generate Hough votes for each point by selecting the centroid of the object they belong to. Another MLP combines the votes' clusters and labels each point with respective semantic class and instance belonging. Compared to their no-vote baseline BoxNet \cite{Qi2019b}, where the 3D boxes are generated directly from PointNet output, VoteNet was able to achieve 18.4 mAP improvement on SUN RGB-D and ScanNet datasets.

ImVoteNet \cite{Qi2020a} improves on VoteNet by including 2D bounding box proposals to generate 3D Hough votes from RGB-D images. The 2D bounding boxes from a Faster R-CNN model are "lifted" to 3D space using the camera parameters and are used to limit the search space for the Voting MLP. Additionally, the semantic and texture modalities from the 2D detector are also used as features in the model. The geometric semantic and texture modalities (cues) are fused together in the final towers which balances the contribution of each signal in the final 3D proposal. Our work adapts the idea of the inverse-projection of the 2D bounding boxes to the point cloud space.

\subsection{3D Shape Completion}
Shape completion techniques are used to infill the missing points in a sparse point cloud based on learned features about the objects. This infilled object point cloud can be used to extract depth of an occluded object. As shown by previous studies \cite{Liu2019a}, CNN cannot consume point clouds directly due to being unordered and irregular and hence point cloud completion networks are used to create an intermediate representation to capture the order and local shape. GRNet, the current state-of-the art dense point cloud completion method uses differential layers called \textit{Gridding} and \textit{Gridding reverse} which capture the structural information during the completion process. \textit{Gridding network} generates a 3D grid $G=<V,K>$, where V is the vertices and and K represents the corresponding values of the set points. The value set is fed to the 3D CNN for segmentation to produce $W'$. Another network \textit{Gridding Reverese} generates a coarser point cloud grid $G'=<V',W'>$ from $G$. From the coarse point cloud $P^c$, features $F^c$ are sampled and processed by an MLP to generate the completed point cloud $P^f$. This method is suitable for our task due to its resilience to a wide range of point cloud resolutions. This removes bias in detecting farthest-low-resolution object points as opposed to closest-high-resolution objects in the 3D pipeline. 

We also explore state-of-the-art RfD-Net \cite{Nie2020} point scene completion method as a possible candidate for point cloud detector for large objects. Their pipeline has 3 parts (1) a 3D detector for seed box proposals (2) a spatial transformer for grouped cuboid proposals and (3) a shape generator for shape completion. While the complete pipeline is proposed as a shape completion technique, we intend to use the output of the spatial transformer, which proposes the aligned 3D cuboids from the point cloud.

\subsection{Robotic Vision Scene Understanding Challenge}
\begin{figure*}[tp]
    \centering
    \begin{minipage}{\textwidth}
    \subfloat[Miniroom]{%
        \includegraphics[width=0.32\textwidth]{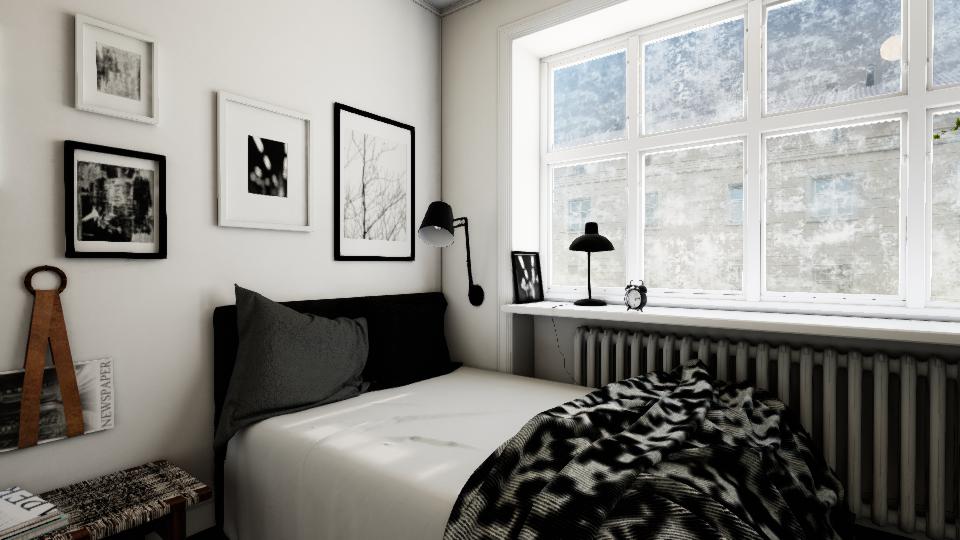}
        \label{fig:miniroom}
    }
    \subfloat[Apartment]{%
        \includegraphics[width=0.32\textwidth]{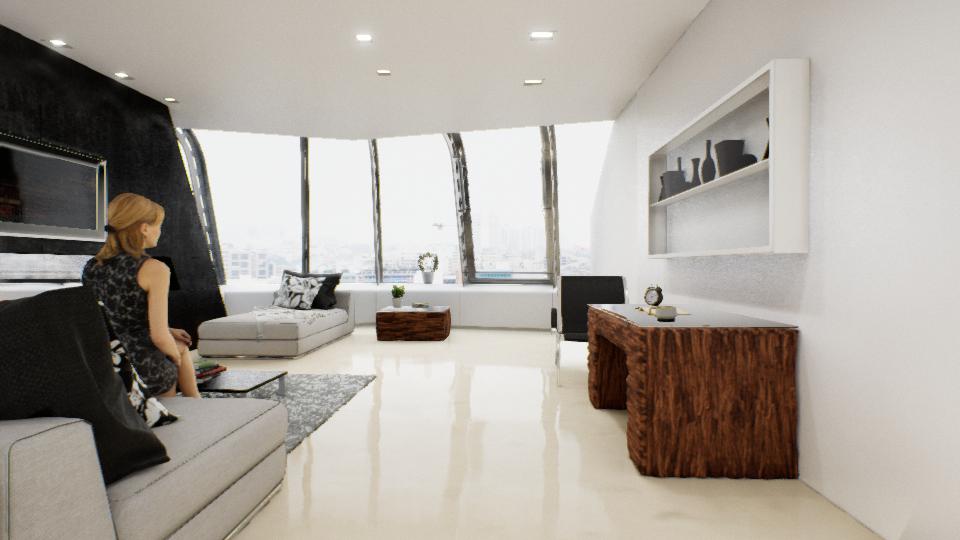}
        \label{fig:apartment}
    }
    \subfloat[Office]{%
        \includegraphics[width=0.32\textwidth]{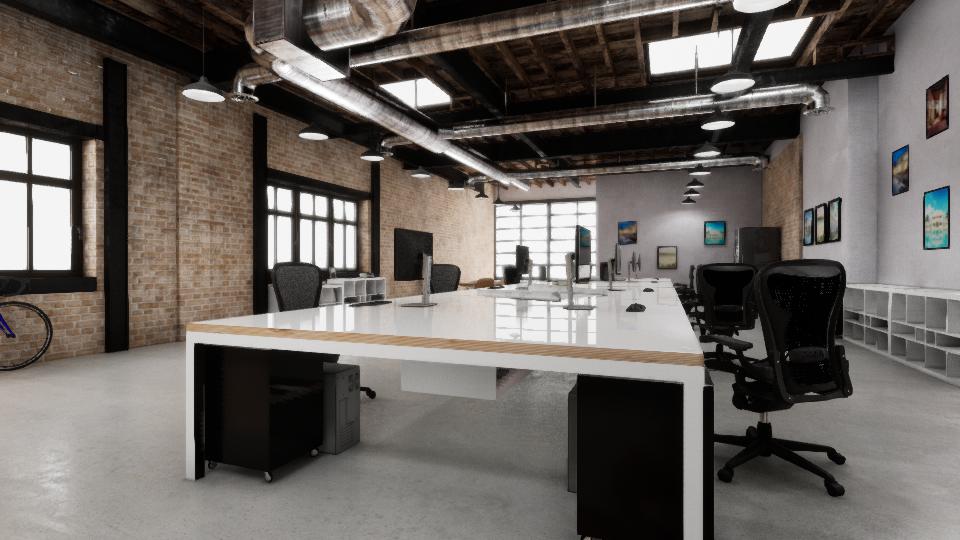}
        \label{fig:office}
    }
    \end{minipage}
    \begin{minipage}{\textwidth}
    \subfloat[House]{%
        \includegraphics[width=0.32\textwidth]{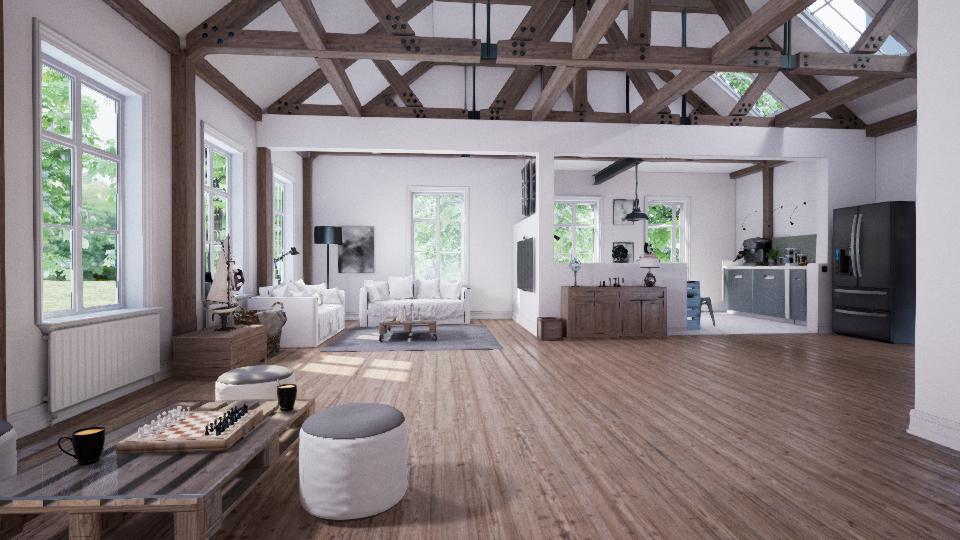}
        \label{fig:house}
    }
    \subfloat[Company (Day)]{%
        \includegraphics[width=0.32\textwidth]{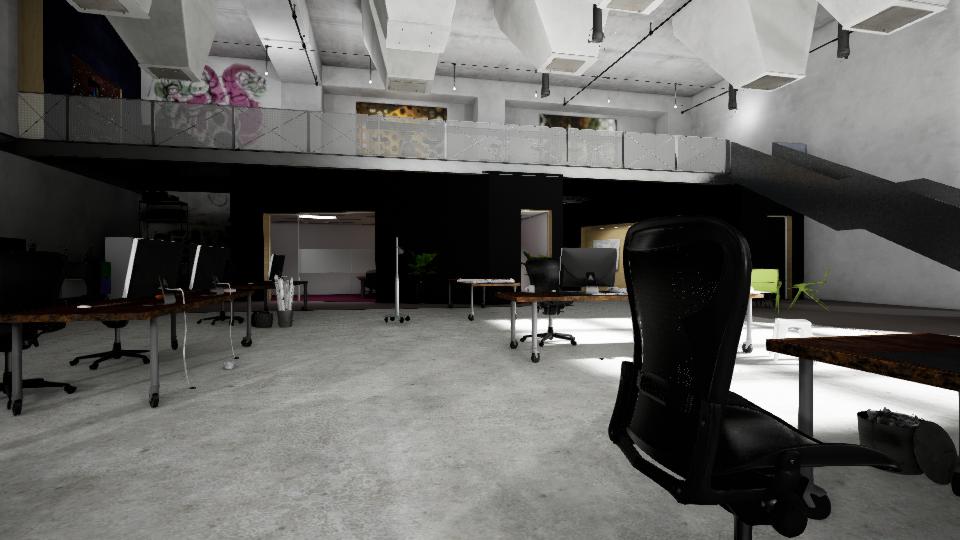}
        \label{fig:companyday}
    }
    \subfloat[Company (Night)]{%
        \includegraphics[width=0.32\textwidth]{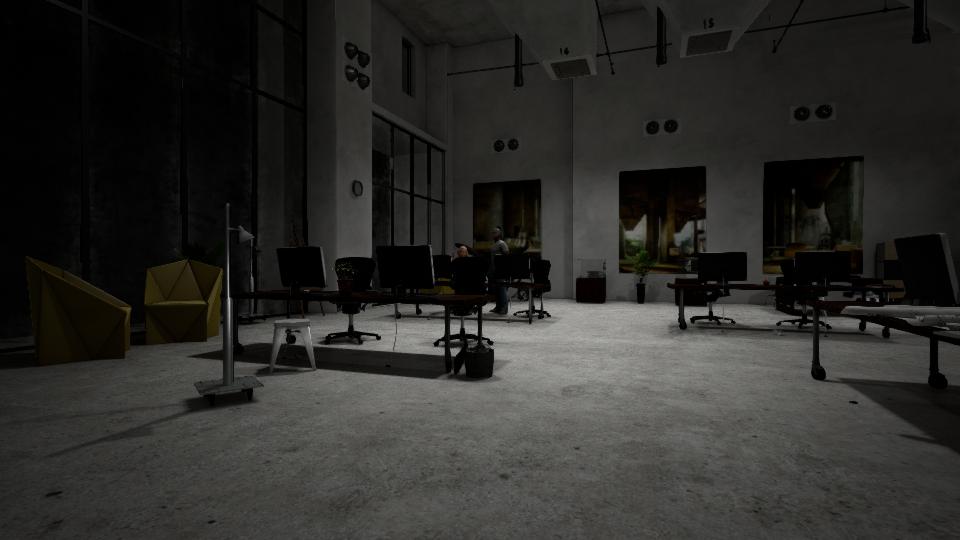}
        \label{fig:companynight}
    }
    \end{minipage}
\caption{Images from the different environments available in the BenchBot simulator. (e) and (f) are examples of time of day variations.}
\label{fig:env-examples}
\end{figure*}

The RVSU Challenge \cite{Hall2020b} is a novel scene understanding task with a high-fidelity photo-realistic simulation, where an active robot agent has to explore indoor environments to detect objects and scene changes. There are 5 base environments namely house, miniroom, apartment, company, and office (see Fig. \ref{fig:env-examples}). The objects are 25 small and large classes selected from COCO dataset at various times of the day (day or night). Each base environments has 5 variations with different objects, time of day, and robot spawn locations. The main challenge is categorised into 2 sub-challenges: Semantic SLAM (SSLAM) and Scene Change detection (SCD). The SSLAM task requires the agent to map objects in the scene in a single run with high accuracy while performing SCD, the agent is required to explore the same base environment twice with variations on the second pass.

Each sub-challenge has 3 difficulty levels (1) Passive actuation with ground truth pose (2) Active control with ground truth pose and (3) Active control and dead reckoning localisation. The first two difficulty levels provide accurate ground truth poses of the robot with land marked-based or free motions, respectively. The highest difficulty allows the robot to be controlled with all degrees-of-freedom but without ground truth poses. The localisation has to be performed by the robot based on the data from its noisy sensors.

In this paper, we test our model on Passive actuation with ground truth pose category. This is to avoid any performance ambiguity introduced by active exploration and localisation implementation.

\subsection{Performance evaluation}\label{sec:performamce-eval}

The Object Map Quality (OMQ) metric \cite{Hall2020b} is used to evaluate the 3D object detection accuracy, which is a probability-based detection metric for evaluating 3D object maps. Unlike 3D IoU, OMQ aims to increase spatial, label and foreground separation qualities. It also penalises false positive ($f_P$) and false negative ($f_N$) detections. OMQ score for a proposed object map $\hat{M}$  given the  ground truth map $M$ is defined as,

\begin{equation}
    OMQ(M, \hat{M}) = \frac{\sum_{i=1}^{N_{TP}} q(i)}{ N_{TP} + N_{FN} + \sum_{j=1}^{N_{FP}} c_{FP}(j) }
\end{equation}
where $N_{TP}$ and $N_{FN}$ are the number of true positives and false negative detections, respectively and $q_{TP}$ is the list of non-zero true-positive quality scores. The false positive cost $c_{FP}$ is list of all false positive label probabilities of object classes. The pairwise object quality (pOQ) score is calculated as geometric mean of a label ($Q_{L}$) and spatial quality ($Q_{Sp}$).

\begin{equation}
pOQ(O_i, \hat O_j) = \sqrt{Q_L(O_i, \hat O_j) \cdot Q_{SP}(O_i, \hat O_j) }
\end{equation}
where $\hat O$ is the proposed detection and $O$ is the ground truth. The label quality ($QL$) is the probability of correct class in the scene. The OMQ metric adapts this metric to evaluate 3D detection quality. 

\section{Methodology}
In this section, we present the complete map generation pipeline used for the scene understanding task. The overall all pipeline architecture is presented first, followed by the detection branch and the used evaluation metrics.  

\subsection{Pipeline}
\begin{figure*}[ht]
    \centering
    \includegraphics[width=\textwidth]{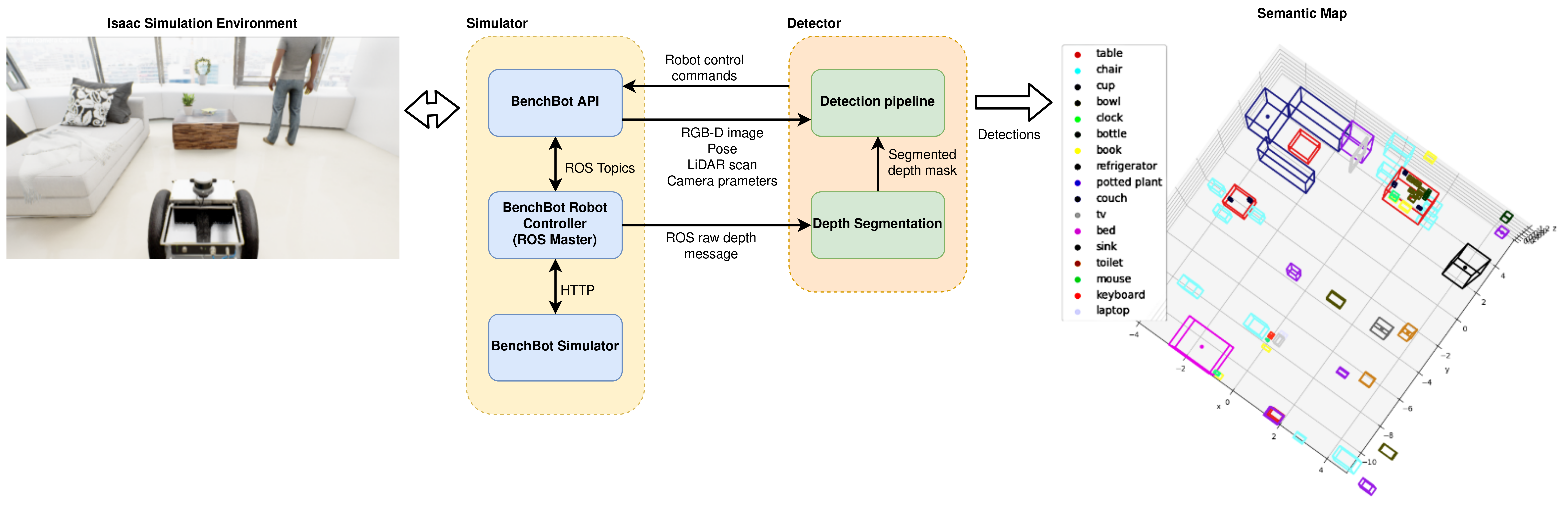}
    \caption{Illustration of the semantic scene understanding pipeline. The Carter robot deployed in the Issac sim \cite{NVIDIA2020} is controlled through the BenchBot API. The RGB-D frames, LiDAR scans, and intrinsic and extrinsic camera parameters are obtained through the API. Additional raw depth images are obtained directly from the ROS master node. The depth segmentation ROS node segments the depth image to obtain surface normals and point cloud "objectness" mask.}
\label{fig:ssu-pipeline}
\end{figure*}
The pipeline (Fig. \ref{fig:ssu-pipeline}) includes the BenchBot robotic environment \cite{Talbot2020a} and a detection system that generates objects map from the sensor outputs. The BenchBot simulator consists of a Isaac sim environment with various room layouts and a robot that can be controlled through an API. While pre-processed data from the on-board sensors can be read using the same API, raw data can be also accessed directly through the Robotic Operating System (ROS) master node in the BenchBot simulator. The detector system consumes the RGB-D frames, robot pose, LiDAR scans, and camera parameters through the Benchbot API to produce the semantic map. This module also sends the control commands such as move, turn, or stop to explore the simulated environment. The depth segmentation node takes raw depth inputs from the BenchBot ROS core to generate a segmented depth mask. After obtaining the final cuboid proposals in the robot reference frame, they are transformed to the map reference frame using the robot pose.    

\subsection{Detection system}
\begin{figure*}[ht]
    \centering
    \includegraphics[width=\textwidth]{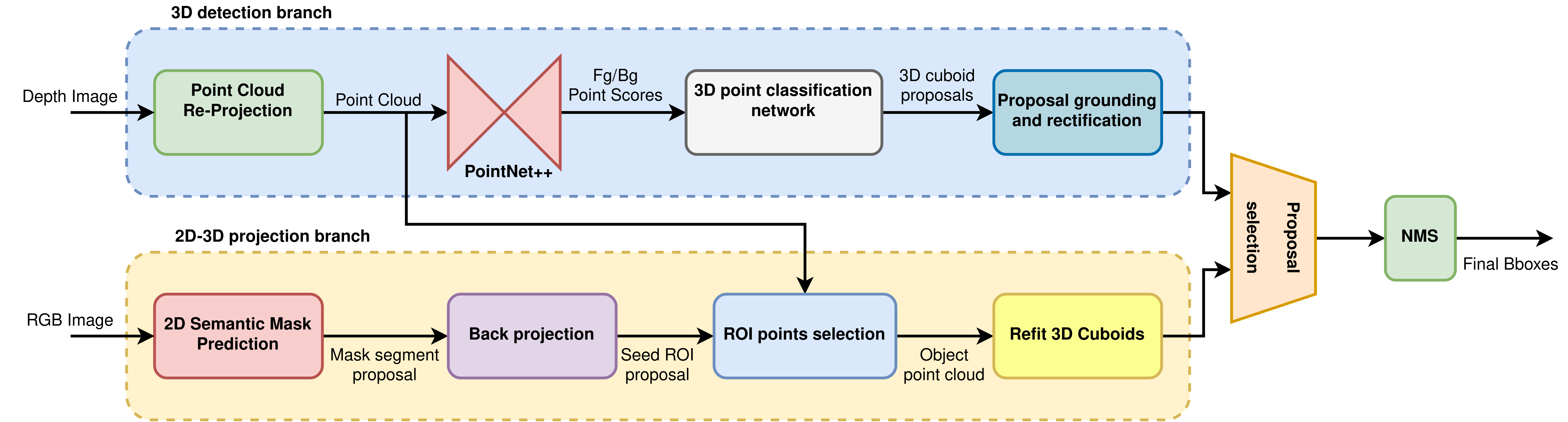}
    \caption{Detection pipeline for the bounding box proposal from RGB-D images. The 2 branches (1) 3D detection for large objects, (2) 2D-3D projection branch for small and medium size objects, independently generate bounding box (Bbox) proposals from the RGB-D input and (3) Proposal fusion and selection.}
\label{fig:detection-section}
\end{figure*}

The detector (Fig \ref{fig:detection-section}) has 3 main sections: (1) 3D detection for large objects and (2) 2D-3D re-projection branch for small and large size objects (3) proposal selection module to fuse the detection from the detection branches based on their confidence and scene category. Branches (1) and (2) independently generate bounding box (Bbox) proposals from the RGB-D input. The PointNet++ backbone scores "objectness" of the points and the 3D point classifier proposes 3D Bboxes from them. This module can be one of VoteNet, ImVoteNet, GRNet or RfDNet. The 2D mask proposed by the Mask RCNN network in the 2D-3D branch is inverse projected to the 3D space using the camera's intrinsic matrix to obtain a seed point cloud of the object. This 3D mask is used to select object points from the point cloud. A bounding box is fitted over the points of each object to obtain the final bounding box. The proposal selection layer uses place recognition to remove false positives.

The 2D-3D inverse projection module uses camera parameters to inverse project the 2D bounding boxes in the image plane to the 3D space using perspective transformation \cite{Hartley2004}. Given the 2D bounding box point $b$ in the image pixel coordinate $p_i = (u,v)$ and the corresponding real world coordinate $P_i =(X, Y, Z)$, the point wise transformation for scale $s$ is 

\begin{equation}
s\begin{pmatrix}
u\\
v\\
1
\end{pmatrix} 
 = C \cdot
 \large\begin{pmatrix}
 X\\
 Y\\
 Z\\
 1
 \end{pmatrix} 
= K \cdot [R_{(3,3)}|t_{(3,1)}] \cdot \large\begin{pmatrix}
 X\\
 Y\\
 Z\\
 1
 \end{pmatrix} 
\end{equation}
where
\begin{equation}
    K = \begin{bmatrix}
    f_x & 0 & x_0\\\
    0 & f_y & y_0 \\
    0 & 0 & 1
    \end{bmatrix}
\end{equation}
is the intrinsic calibration matrix for the perspective camera with principal and focal points at $(x_0, y_0)$ and $(f_x, f_y)$, respectively. $C$ is a perspective transform $C: (X,Y,Z) \in R^3 \xrightarrow[]{} (u,v) \in R^2$, and $R$ and $t$ are the camera rotation and translation vectors, respectively. While recovering the actual depth $Z$ by $C^{-1}$ is impossible analytically, using the sparse depth values from the RGB-D frame will limit the search space to a relatively small region. We use this projected frustum bounding box as a seed 3D proposal for further rectification. We use the depth segmentation algorithm from \cite{Furrer2018} to infill and shape-complete the seed point clouds. An axis aligned bounding box is fitted on these rectified points to obtain the final cuboid proposal $B_i$. The overall algorithm is detailed in Algorithm \ref{algo:pointcloud-rectification}. Here $C^{+}$ is the Moore-Penrose pseudo inverse \cite{Campbell2009} of the projection matrix $C$.

\begin{algorithm}[]
\DontPrintSemicolon
\caption{Rectify And Project 2D bounding box}
\KwIn{$b$, $P_c$ $K$, $R$, $t$}
\KwOut{Rectified projected bounding boxes $B$}
B = $\emptyset$\;
\For{ $b_i \in b$ }{
    $C \xleftarrow[]{} K [R | t]$ \;
    $B_i \xleftarrow[]{}  C^{+} b_i$ \;
    $p_{CP} \xleftarrow[]{} crop(B_i, P_c)$ \;
    $p_{SC} \xleftarrow[]{} depth\_segmentation(p_{CP})$ \;
    $B_i \xleftarrow[]{} fit\_axis\_aligned\_box(p_{SC})$ \;
}
\Return{B}\;
\label{algo:pointcloud-rectification}
\end{algorithm}

\subsection{Proposal fusion and selection}
In order to improve any likely low class accuracy from the 3D detection branch, we use the 2D bounding box proposals to \textit{confirm} the object presence in the scene.  For this the IoU of the refitted cuboids from the 2D and 3D branches are calculated. An $IoU > 0.5$ with same class labels suggests a significant overlap in the detection and thus \textit{confirms} the object class from the 3D detection branch. This removes false positive detections from the 3D pipeline. Additionally, the detections of large objects from the 3D branch and small objects from the 2D branch are both used in the final detection result. Next, we apply the 3D non-maxima suppression (NMS) method \cite{Kumar2021} to remove any duplicate detections.

\section{Experiments}
\subsection{Dataset}
The experiments in this study are conducted on BenchBot \cite{Talbot2020a}, a simulation-based indoor environment that has photo-realistic everyday scenes in which a robot is deployed and navigated. Like a real robot, the simulated odometry, LiDAR, RGB, and depth camera sensor data are accessible through BenchBot API. As the environment is built on a full-fledged game engine, the physics and dynamics are near-realistic. The motion of the robot is controlled by sending commands through same API. 

We use SUN RGB-D \cite{slx15} and ScanNet \cite{dcshfn17} datasets for training and benchmarking our 3D pipeline. The SUNRGB-D dataset consists of 10K RGB-D images with 64K 3D amodal oriented bounding boxes with their camera parameters captured from various indoor single view scenes. Out of 800 categories, the objects that are relevant to the SSU challenge are identified and selected. Due to the difference in category naming, a mapping is maintained between the BenchBot and SUN RGB-D classes (i.e. sofa vs couch). We augment this selected dataset with objects from ScanNet to include other objects in BenchBot classes. The ScanNet dataset consists of 2.5M RGB-D images with proposal cuboid and semantic segmentation labels for each object in 1513 scenes. The depth images from the augmented dataset are converted to point cloud data (PCD) using the camera's intrinsic and projection matrices available with each dataset. This augmented RGB+PCD dataset is used for training and evaluating our 3D pipeline.

\subsection{Training}
We select PointNet++ \cite{Qi2017} as our backbone network for its state-of-the-art performance in direct point cloud detection. PointNet++ is trained on our augmented dataset with 16 images per batch. This pretrained backbone network is used for evaluating VoteNet, ImVoteNet and RfDNet models. Each network is trained end-to-end by freezing the backbone PointNet++ layer for comparison.
Training of all models is conducted on a machine with Intel Core i7 3.70 GHz CPU and NVIDIA Quadro RTX Quadro 8000 GPU. 

\subsection{Testing}
The pipeline is run on the BenchBot evaluation benchmark using OMQ metric detailed in Section \ref{sec:performamce-eval}. The chosen testing environments are 2 variations of office, apartment, and company. In order to evaluate the computational cost for each model, the experiment was run 3 times on each model and the run times were recorded. The average detection time per object is calculated by,

\begin{equation}
    \mu_T = \frac{Total\ number\ of\ detections}{Total\ time\ taken\ per\ run}
\end{equation}

\section{Results and Discussion}

\begin{table*}[!ht]
\renewcommand{\arraystretch}{1.3}
\caption{Comparison of the pipeline OMQ scores with various proposal cuboid proposal methods}
\centering
\begin{tabular}{l|c|c|c|c|c}
 \hline\hline
 \bfseries Method                 & \bfseries OMQ           & \bfseries Avg. Pairwise & \bfseries Avg. label    & \bfseries Avg. spatial  & \bfseries Avg. fp quality \\ \hline\hline
 VoteNet \cite{Qi2019b} (First Baseline)                                    & 0.11          & 0.63          & 0.92          & 0.44          & 0.35            \\
ImVoteNet  \cite{Qi2020a}                                  & 0.19          & 0.55          & 0.94          & 0.35          & 0.26            \\ 
VoteNet\cite{Qi2019b} + Mask Scoring R-CNN \cite{Huang2019c}                             & 0.25          & 0.62          & 0.1          & 0.42          & 0.03            \\ 
GRNet \cite{Xie2020} + Mask Scoring R-CNN \cite{Huang2019c}                            & 0.33          & 0.65          & 0.98          & 0.48          & 0.11            \\ 
DepthSeg \cite{Furrer2018} + BMask R-CNN \cite{Cheng2020}  (Second Baseline)                        & 0.44          & 0.68          & 1.00          & 0.51          & 0.28           \\  \hline
Ours (RfDNet \cite{Nie2020} + Mask Scoring R-CNN \cite{Huang2019c})                    & 0.46          & 0.70          & 1.00          & 0.55          & 0.32   \\         
\textbf{Ours} (RfDNet \cite{Nie2020} + Mask Scoring R-CNN \cite{Huang2019c} + PropSel) & \textbf{0.49} & \textbf{0.71} & \textbf{1.00} & \textbf{0.55} & \textbf{0.37}   \\ \hline\hline
\end{tabular}
\label{tab:result-comp}
\end{table*}
We evaluated our pipeline on RVSU challenge test scenes (office, apartment and company) as per the competition requirement (see Table \ref{tab:result-comp}). While the ImVoteNet model improved the overall OMQ, it is sub-optimal in its spatial and false positive quality compared to the first baseline (VoteNet). Adding Mask projection on VoteNet detections improved results over ImVoteNet by 0.6 OMQ. To evaluate the performance of the 3D shape completion method, we tested GRNet with the same point cloud resolution of ImVoteNet and VoteNet. The OMQ score showed significant improvement by a factor of 0.8 OMQ. The second baseline (DepthSeg [20] + BMask R-CNN) had a higher OMQ owing to better pairwise (0.68), label (1.00) and spatial (0.51) performances. While it didn't have a dedicated 3D detection module, the baseline used a combination of 2D projection and 3D segmentation methods. Inspired by the success of this method, we fused the outputs of RfD-Net, a complete 3D detector, and Mask Scoring RCNN, a 2D segmentation model. The OMQ score improved on the second baseline by 0.2 OMQ. To improve our pipeline performance for scene variations (day, night, and shade) we applied the proposal selector on this output. We were able to achieve a final OMQ of 0.49. Additionally, a comparison of the detection speed of each method (Table \ref{tab:speed-comp}) shows that the 3D-only methods have lower average detection times compared to combined 2D and 3D strategies as be the expectation. Our method successfully produced a higher OMQ score without a significant loss in computation time per detection. 

The VoteNet model with Mask Scoring R-CNN greatly improved the label and spatial quality by separating the 3D and 2D detection. While the objects in dark scenes in the test environments were missed by the 2D detector, VoteNet was able to detect them with high confidence. The Shape completion method GRNet improved the spatial quality but aggravated the false positive quality. A qualitative analysis shows a significant amount of false positive detections from the partial depths (e.g. a portion of a couch detected as bathtub).  The baseline model using VoteNet missed small objects such as monitor, bottle, potted plant, etc. On the other hand, it performed well when detecting large objects such as a bed or a couch. ImVoteNet was able to improve the label quality by using the 2D detection features on the RGB input. Incidentally, the pairwise and spatial quality dropped lower than the initial VoteNet baseline as the under-confident 2D detections suppress the 3D detections as reported by \cite{Qi2020a}. 

In the second baseline, which was propose by the challenge authors \cite{Hall2020b}, the output of the depth segmentation method \cite{Furrer2018} for cuboid proposal was fused with the BMask R-CNN  for mask prediction. While this baseline was able to achieve a higher OMQ than dedicated 3D methods, it used intuition-based post-processing methods to remove false predictions. The post-processing methods expected the larger object to be on ground level and smaller objects to be limited by size. Special cases such as a wall mounted television and toilets are counter examples to this expectation. Our method used the RfD-Net scene construction method as the 3D detector in the proposed pipeline and Mask Scoring R-CNN for 2D mask prediction. Mask Scoring R-CNN is better for projection, as the BMask R-CNN includes occluding object boundaries in the mask causing its projection to overestimate the boundaries. Our proposal selection (PropSel) removed false positive 3D detections by selecting proposals that have high 3D IoU between 3D and projected 2D proposals. This resulted in obtaining the highest final OMQ among the evaluated methods.

\subsubsection{Limitations}
Detecting unusually large objects like conference tables (like Fig. \ref{fig:office}) are still a challenge to our method. Although we were able to merge multiple detections of conference table from different angles to cover a significant portion of the table, there is still room for improvement in this front. One possible solution is to register the same object in different angles and fuse them using Structure from Motion (SfM).As future work we intent to extend this pipeline to address this problem. 

\begin{table*}[!h]
\renewcommand{\arraystretch}{1.5}
\caption{Comparison of the average time per detection for evaluated methods in Miniroom:1 environment}
\centering
\begin{tabular}{l|c}
 \hline\hline
 \bfseries Method & \bfseries Avg. detection time per object $\mu_T$ \\ \hline\hline
 VoteNet \cite{Qi2019b} (First Baseline)  &  1.19s \\
ImVoteNet  \cite{Qi2020a} & 1.22s \\ 
VoteNet\cite{Qi2019b} + Mask Scoring R-CNN \cite{Huang2019c} & 1.24s \\ 
GRNet \cite{Xie2020} + Mask Scoring R-CNN \cite{Huang2019c} &  2.46s \\ 
DepthSeg \cite{Furrer2018} + BMask R-CNN \cite{Cheng2020}  (Second Baseline) &  1.87s \\  \hline
Ours (RfDNet \cite{Nie2020} + Mask Scoring R-CNN \cite{Huang2019c}) & 1.81s   \\         
\textbf{Ours} (RfDNet \cite{Nie2020} + Mask Scoring R-CNN \cite{Huang2019c} + PropSel) & 1.90s  \\ \hline\hline
\end{tabular}
\label{tab:speed-comp}
\end{table*}

\section{Conclusion}

Semantic scene understanding is a challenging and significant task in Embodied AI. In this work, we were able to demonstrate the successful fusion of 2D and 3D detection techniques to improve the evaluated performance on the semantic scene understanding challenge. We tackled this challenge by building a pipeline that is integrated with a photo-realistic robotic platform. We compared our proposed pipeline to various benchmarks and proposed methods. The results of performance evaluation demonstrated that 2D methods can improve 3D detection results significantly by reducing the false positive detections. In future work, we plan to extend on this study to improve embodied vision-language navigation tasks.






\bibliographystyle{IEEEtran}
\bibliography{bare_conf.bib}
%



\end{document}